\documentclass[11pt]{article}

\usepackage[final]{acl}

\usepackage{times}
\usepackage{latexsym}
\usepackage{adjustbox}

\usepackage[T1]{fontenc}

\usepackage{microtype}

\usepackage{inconsolata}

\usepackage{graphicx}
\usepackage{amsmath}
\usepackage{booktabs}
\usepackage{tcolorbox}
\usepackage{amssymb}
\title{Preference Estimation via Opponent Modeling in Multi-Agent Negotiation}

\author{
  \textbf{Yuta Konishi}\textsuperscript{1},
  \textbf{Kento Yamamoto}\textsuperscript{2},
  \textbf{Eisuke Sonomoto}\textsuperscript{2},
  \textbf{Rikuho Takeda}\textsuperscript{2}, \\
  \textbf{Ryo Furukawa}\textsuperscript{2},
  \textbf{Yusuke Muraki}\textsuperscript{2},
  \textbf{Takafumi Shimizu}\textsuperscript{1},
  \textbf{Kazuma Fukumura}\textsuperscript{1}, \\
\textbf{Yuya Kanemoto}\textsuperscript{2,$\dagger$},
  \textbf{Takayuki Ito}\textsuperscript{1},
  \textbf{Shiyao Ding}\textsuperscript{1,$\dagger$} \\ 
  \textsuperscript{1}Graduate School of Informatics, Kyoto University, Kyoto, Japan\\
  \textsuperscript{2}Accenture Japan Ltd, Tokyo, Japan\\
  \small{
    \textsuperscript{$\dagger$}
    \texttt{yuya.kanemoto@accenture.com},
    \texttt{ding@i.kyoto-u.ac.jp}
  }
}

\begin{document}
\maketitle

\begin{abstract}
Automated negotiation in complex, multi-party and multi-issue settings critically depends on accurate opponent modeling. However, conventional numerical-only approaches fail to capture the qualitative information embedded in natural language interactions, resulting in unstable and incomplete preference estimation. Although Large Language Models (LLMs) enable rich semantic understanding of utterances, it remains challenging to quantitatively incorporate such information into a consistent opponent modeling. 
To tackle this issue, we propose a novel preference estimation method integrating natural language information into a structured Bayesian opponent modeling framework. Our approach leverages LLMs to extract qualitative cues from utterances and converts them into probabilistic formats for dynamic belief tracking. Experimental results on a multi-party benchmark demonstrate that our framework improves the full agreement rate and preference estimation accuracy by integrating probabilistic reasoning with natural language understanding.
\end{abstract}

\section{Introduction}
In modern society, automated negotiation is a pivotal technology for conflict resolution and efficient consensus-building among diverse stakeholders \citep{memon2025systematicmapping, bagga2021anegma}.
Historically, the field has matured through integrated development environments like the General Environment for Negotiation with Intelligent 
multi-purpose Usage Simulation (GENIUS)~\citep{genius} and international competitions such as the Automated Negotiating Agents Competition (ANAC)~\citep{anac}. 
A significant milestone was the BOA architecture~\citep{boa}, which standardized negotiating agents into three decoupled components: the \textit{bidding strategy}, \textit{opponent model}, and \textit{acceptance strategy}. 
Within multi-party, multi-issue settings, opponent modeling remains essential for strategic decision-making \citep{baarslag2016learning}. Traditionally, these models have evolved through Bayesian learning \citep{zeng1998bayesian, opponent-modeling} and reinforcement learning \citep{he2016deep}, primarily estimating utility functions from numerical proposal histories.
However, numerical-only methods struggle to capture qualitative contexts, leading to unstable estimation under high information uncertainty \citep{baarslag2016learning}.

To address these limitations, integrating Large Language Models (LLMs) into negotiation and decision-making frameworks has gained traction \citep{cooperation, fu2023improving}. LLMs possess sophisticated capabilities for context understanding and Theory of Mind (ToM) \citep{Kosinski2024tom, negotiationtom}, enabling the extraction of qualitative preference signals typically lost in conventional models. Nevertheless, directly applying reasoning techniques like Chain-of-Thought (CoT)~\citep{cot}, Tree of Thoughts (ToT)~\citep{tot}, or Multi-Agent Debate (MAD)~\citep{mad} to LLM-based agents reveals new challenges: a lack of strategic consistency during prolonged negotiations \citep{negotiationtom}, fragile generalization across different problem settings \citep{zhao-etal-2025-large}, and an exponential increase in inference complexity as the amount of available information grows \citep{cooperation}.
In addition, prior work on natural language negotiation using LLMs \citep{tombench, negotiationtom} has largely focused on intent inference in static or short-horizon evaluation settings, where strategic dynamics are limited. Such approaches often lack a formal mechanism for belief updating over time, thereby hampering stable preference tracking in dynamic negotiation scenarios.

\begin{figure*}[t]
  \includegraphics[width=\linewidth]{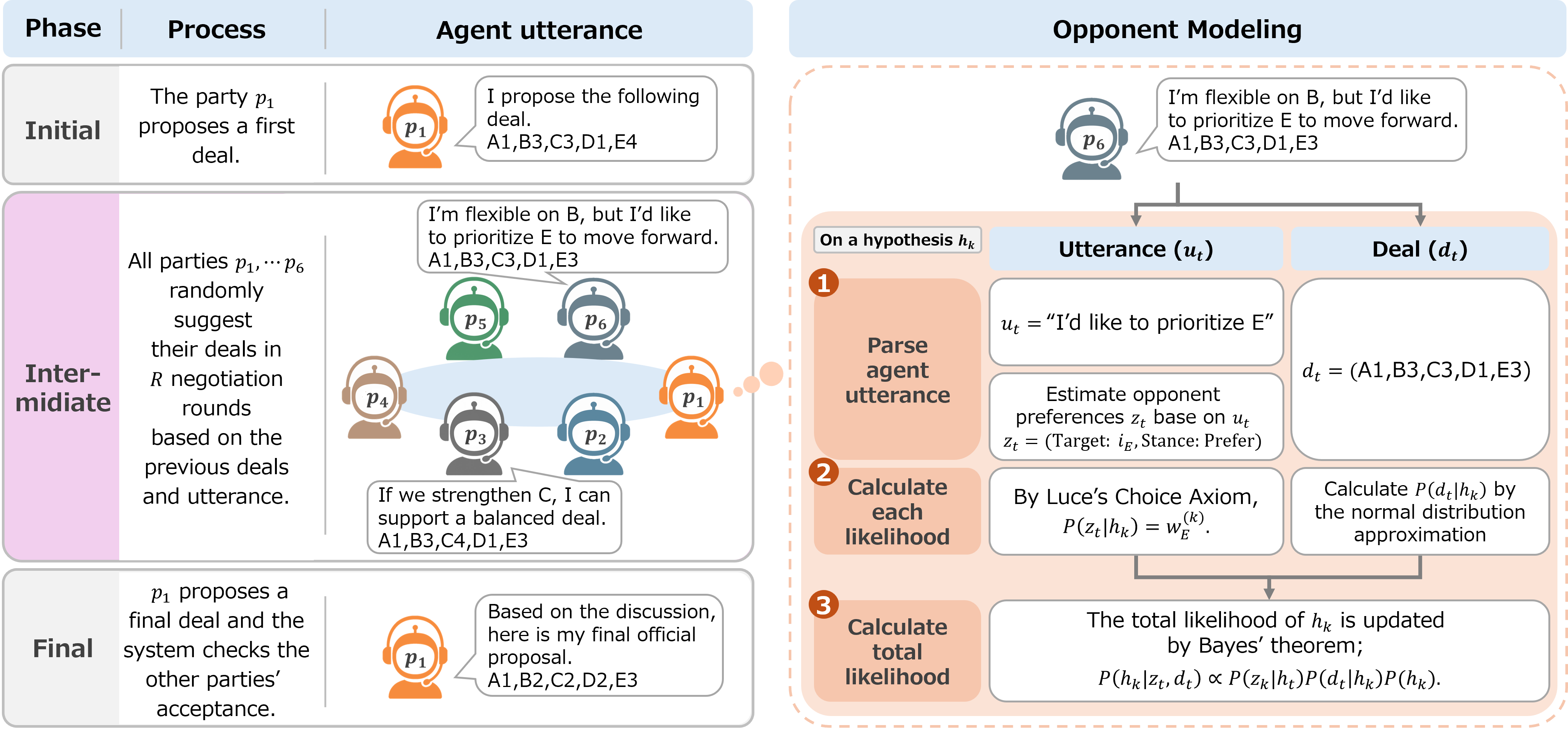}
  \caption{
    Overview of the negotiation flow (left) and the proposed Bayesian opponent modeling process (right).
    The left panel shows the three phases: initial proposal by $p_1$, intermediate rounds with deal and utterance exchanges
    among agents $p_1 \dots p_6$, and the final proposal by $p_1$. 
    Each deal at round $t$ is represented as $d_t = (o_t^1, \dots, o_t^M)$, where $o_t^m$ denotes the selected option for issue $i_m$.
    In this example, issues are denoted by capital letters and options by indices. For instance, the initial proposal $d_1 = (A1, B3, C3, D1, E4)$ corresponds to
    $d_1 = (o_1^1 = A1, o_1^2 = B3, o_1^3 = C3, o_1^4 = D1, o_1^5 = E4)$
meaning that option 1 is selected for issue $A$, option 3 for issue $B$, and so on.
    The right panel illustrates $p_1$’s internal modeling for a hypothesis $h_k$, consisting of:
    (1) parsing the opponent’s utterance $u_t$ to estimate preference signals $z_t$;
    (2) calculating the likelihood for $z_t$ (via Luce’s Choice Axiom) and the likelihood of the proposed deal $d_t$
    (via normal distribution approximation); and
    (3) updating the posterior probability of $h_k$ through Bayesian fusion.
    }
  \label{fig:negotiation_and_opponentmodeling}
\end{figure*}

To address these challenges, we propose a novel preference estimation method that integrates natural language signals from dialogue into a structured Bayesian framework. Our approach utilizes LLMs to extract qualitative cues and subsequently converts these cues into a format compatible with probabilistic models for dynamic belief tracking. 

Our main contributions are summarized as follows. First, we propose an integrated framework that complements qualitative intent extraction via LLMs with quantitative preference estimation through Bayesian inference. Second, we demonstrate that the proposed method achieves superior preference estimation performance in complex multi-party scenarios compared to baselines relying solely on numerical data or direct LLM inference. Finally, we show that our framework improves agreement rates even under high uncertainty, thereby facilitating more effective autonomous negotiation.

\section{Problem Formulation}
We adopt the scorable negotiation framework proposed in \cite{cooperation}. 
Let $P = \{p_1, \dots, p_N\}$ denote the set of parties with $p_n \in P$, and $I = \{i_1, \dots, i_M\}$ denote the set of issues with $i_m \in I$, where each issue $i_m$ has a finite option set $O_m = \{o^m_1, \dots, o^m_{K_m}\}$ with $o^m_k \in O_m$. 
A deal $d_t$ proposed at round $t$ is defined as a tuple of options $d_t = (o^1_t, \dots, o^M_t)$, where each $o^m_t \in O_m$ is selected for the corresponding issue $i_m$.

Each party $p_n$ holds a private score function $s_n^m : O_m \to \mathbb{R}$ for each issue $i_m \in I$, and the utility of a deal $d_t$ for party $p_n$ is defined as the sum of these scores:
\begin{equation}
U_n(d_t) = \sum_{m=1}^{M} s_n^m(o^m_t).
\end{equation}
Upon reaching an agreement, each party receives the utility $U_n(d_t)$. Otherwise, each party receives their Best Alternative to a Negotiated Agreement (BATNA), represented by a private reservation threshold $\tau_{p_n} \in \mathbb{R}$.

The left panel of Figure~\ref{fig:negotiation_and_opponentmodeling} illustrates the flow of the negotiation.
The negotiation lasts for up to $T$ rounds. In each round $t$, a designated party proposes a deal $d_t$ and a natural language utterance $u_t$, without revealing score functions; parties infer other parties’ preferences from the history of $d_t$ and $u_t$. The success of the negotiation is determined by the deal $d_T$ in the final round. An agreement is reached if and only if at least a minimum required number of parties, including all veto holders, satisfy $U_n(d_T) > \tau_{p_n}$.

\section{Bayesian Preference Estimation Method}
\label{sec:method}

In this section, we describe our method for explicitly estimating opponents’ preferences at each negotiation round. Our approach builds on the Bayesian opponent modeling framework established by \citet{opponent-modeling}, extending it to integrate natural language information using an LLM. The right panel of Figure~\ref{fig:negotiation_and_opponentmodeling} illustrates the specific mechanism of our proposed opponent modeling framework. 

\subsection{Model Representation and Hypothesis Space}
First, we define the representation of the opponent's strategy and the space of possible preferences. We represent an opponent's preference model using two components: an issue-weight vector $\mathbf{w} = [w_1, \dots, w_M]$, which captures the relative importance of each issue, and a set of evaluation functions $\mathbf{v} = [v_1, \dots, v_M]$, where each $v_m$ specifies the preference shape over the options of issue $i_m$. Based on these components, the agent maintains a finite hypothesis space over possible opponent preferences.

% \noindent \textbf{Hypothesis Space} \\
\paragraph{Hypothesis Space.}
To estimate the opponent’s score function, the agent maintains a finite set of candidate hypotheses $H = \{h_1, \dots, h_K\}$. Each hypothesis $h_k \in H$ represents a specific combination of a weight vector $\mathbf{w}^{(k)} = [w_1^{(k)}, \dots, w_M^{(k)}]$, denoting the relative importance of each issue, and a vector of evaluation functions $\mathbf{v}^{(k)} = [v_1^{(k)}, \dots, v_M^{(k)}]$, representing the preference shapes for each issue. 

% \noindent \textbf{Estimated Utility Function} \\
\paragraph{Estimated Utility Function.}
Under a given hypothesis $h_k$, the estimated utility $\hat{U}(d_t; h_k)$ of a deal $d_t$ is modeled as an additive utility function. It is calculated as the weighted sum of the evaluation functions:
\begin{equation}
\hat{U}(d_t; h_k) = \sum_{m=1}^{M} w_m^{(k)} \cdot v_m^{(k)}(o_t^m).
\end{equation}

\paragraph{Likelihood Based on Numerical Offers.}
Assuming the opponent follows a concession-based strategy, we define the likelihood $P(d_t \mid h_k)$ of observing a deal $d_t$ under hypothesis $h_k$. This is based on the proximity between the estimated utility $\hat{U}(d_t; h_k)$ and $u'(t)$:
\begin{equation}
P(d_t \mid h_k) \propto \exp\left( - \frac{(\hat{U}(d_t; h_k) - u'(t))^2}{2\sigma^2} \right).
\end{equation}
Here, $u'(t)$ denotes the opponent's assumed target utility at round $t$, reflecting a concession-based aspiration level over time. Although we adopt a concession-style strategy here as a standard baseline for the offer-based likelihood, this component is modular and can be replaced with other behavioral models without changing the linguistic likelihood or the Bayesian fusion rule. 

In addition, despite the factorial growth of the hypothesis space as the number of issues increases, prior work~\citep{opponent-modeling} has proposed scalable approximations for this class of Bayesian opponent modeling, which can be incorporated into our framework without changing the Bayesian update itself.

\subsection{Linguistic Likelihood Estimation via LLM}
We describe how linguistic utterances are converted into probabilities over opponent preferences. As a complement to the likelihood from observed deals in Eq.~(3), we define a linguistic likelihood over opponent preferences based on utterances.

\paragraph{Signal Extraction via LLM.}
We use an LLM to parse an utterance $u_t$ into a structured signal $z_t$. Each signal $z_t$ is represented by the following two attributes:

\begin{itemize}
    \item \textbf{Target}: the issue or option referred to by the signal. A target can take one of four forms: (i) a single issue, (ii) a comparison between two issues, (iii) a single option, or (iv) a comparison between two options.
    \item \textbf{Stance}: the attitude toward the target, such as ``prefer'' or ``oppose.'' Together, these attributes allow the agent to convert qualitative information---for example, ``Issue $i_1$ is important'' or ``Option $o^1_{1}$ is preferable to $o^1_{2}$''---into a form suitable for probabilistic computation.
\end{itemize}

In the current formulation, we assume that such linguistic signals are broadly truthful. This assumption could be relaxed in future work by introducing a reliability parameter that dynamically controls the contribution of linguistic evidence based on its consistency with observed offers and dialogue over the course of the negotiation.

% \noindent \textbf{Likelihood Calculation based on Luce's Axiom} \\
\paragraph{Likelihood Calculation based on Luce's Axiom.}
To quantify the likelihood $P(z_t \mid h_k)$, we apply Luce's Choice Axiom. For instance, the probability of observing a signal that indicates a preference for issue $i_x$ is defined as:
\begin{equation}
P(z_t \in \mathcal{Z}_{i_x, \text{pref}} \mid h_k) = \frac{w_x^{(k)}}{\sum_{m=1}^{M} w_m^{(k)}},
\end{equation}
where $\mathcal{Z}_{i_x, \text{pref}}$ denotes the set of signals representing a ``prefer'' stance toward issue $i_x$. 
Similarly, likelihoods for comparison or opposition are calculated based on the relative ratios of components within $\mathbf{w}^{(k)}$ and $\mathbf{v}^{(k)}$ for each hypothesis $h_k$.

\subsection{Preference Update via Multimodal Observations}
We now integrate the numerical-offer likelihood in Eq.~(3) and the linguistic likelihood in Eq.~(4) into a unified Bayesian update rule. 
For simplicity, we adopt a Naive Bayes assumption under which the numerical offer $d_t$ and the linguistic signal $z_t$ are conditionally independent given a hypothesis $h_k$. 
Under this assumption, the posterior distribution over hypotheses is updated as follows:

\begin{equation}
\begin{split}
P(h_k | d_t, z_t) &= \frac{P(d_t, z_t | h_k)P(h_k)}{P(d_t, z_t)}, \\
% &= \frac{P(d_t | h_k)P(z_t | h_k)P(h_k)}{\sum_{h' \in H} P(d_t, z_t | h')P(h')} \\
&\propto P(d_t | h_k)P(z_t | h_k)P(h_k).
\end{split}
\end{equation}

Here, $P(d_t | h_k)$ denotes the likelihood of the numerical offer under hypothesis $h_k$, $P(z_t | h_k)$ denotes the likelihood of the linguistic signal, and $P(h_k)$ denotes the prior probability of $h_k$. 
By computing these terms for each hypothesis and normalizing the resulting scores, the agent sequentially updates its posterior belief over the opponent's preferences, with the linguistic likelihood $P(z_t | h_k)$ favoring hypotheses consistent with the opponent's stated preferences.

\section{Experiments}
\subsection{Experimental Setup}
We evaluate our method using the multi-agent negotiation environment proposed by \citet{cooperation}.

\paragraph{Negotiation Scenario.}
We evaluate our method on a negotiation scenario involving the construction of a sports facility with $N=6$ stakeholders, including two veto holders ($p_1, p_2$), and $M=5$ issues. We chose this scenario because it provides a challenging testbed in which the quality of preference estimation has a direct impact on agreement outcomes. In all experiments, we set $T=24$. The scenario is characterized by diverse preferences among the parties, making consensus building highly difficult: among all 720 possible deals, only 2.9\% (21 deals) satisfy the reservation thresholds $\tau_{p_n}$ for at least five parties including the veto holders, and only 0.4\% (3 deals) satisfy the thresholds for all six parties.

\paragraph{Methods.}
We compare the proposed method with three baselines to assess the effectiveness of Bayesian preference estimation in multi-agent negotiation.

\begin{itemize}
    \item \textbf{Proposed}: In our proposed method, we evaluate two configurations: \textit{p1}, where only the leader $p_1$ performs preference estimation, and \textit{all}, where all agents perform mutual preference estimation.
    \item \textbf{Base-LLM}: The original implementation, in which agents negotiate solely based on prompting, without explicit preference estimation.
    \item \textbf{Baseline Opponent Modeling (Base-OM)}: A conventional Bayesian approach that estimates preferences using only the history of deals $d_t$.
    \item \textbf{LLM Preference Estimation (LLM-PE)}: A method in which an LLM directly infers the numerical values of the opponents' score functions $s_n^m$, without using a structured Bayesian framework.
\end{itemize}

\paragraph{LLM Configuration.}
For all methods, GPT-4.1 was used as the underlying model. 
Since our goal is to evaluate the proposed Bayesian preference estimation algorithm rather than compare foundation models, using a single sufficiently capable LLM is adequate for this study. 
% Although the framework is model-agnostic, evaluation with other frontier LLMs remains future work.

\subsection{Evaluation Metrics}
To assess the negotiation outcomes in the previously described sparse agreement space, we calculate the mean values of the following three metrics across 500 independent negotiation trials for each method. Given the stochastic nature of LLM-based interactions, this extensive number of trials ensures that the reported averages represent stable performance trends and mitigate the influence of individual trial variability.

\begin{itemize}
    \item \textbf{Full Agreement Rate (FAR)}: The ratio of trials where all six parties reached a consensus.
    \item \textbf{Partial Agreement Rate (PAR)}: The ratio of trials where an agreement was reached by at least five parties including the veto holders in the final round $d_T$.
    \item \textbf{Latent Agreement Rate (LAR)}: The ratio of trials where at least one valid deal was proposed during the $T$-round process.
\end{itemize}
Furthermore, we measure the Mean Squared Error (MSE) between the estimated score functions and the true score functions $s_n^m$ to evaluate estimation accuracy.

\begin{table}[t]
  \centering
  % \small
  % \setlength{\tabcolsep}{3pt}
  \caption{
      Comparison of negotiation outcomes across different methods. 
      The performance is evaluated based on the Full Agreement Rate (FAR), Partial Agreement Rate (PAR), and Latent Agreement Rate (LAR). 
      Bold values indicate the best performance for each metric.
      Over 500 runs, the standard deviations were at most $0.02$ across all methods and metrics.
  }
  \label{tab:outcomes_game_a}
  \begin{tabular}{lccc}
    \toprule
    \textbf{Method} & \textbf{FAR} & \textbf{PAR} & \textbf{LAR} \\
    \midrule
    \textbf{Proposed (p1)} & 0.46 & 0.78 & 0.96 \\
    \textbf{Proposed (all)} & \textbf{0.62} & 0.89  & 0.98 \\
    \textbf{Base-LLM} & 0.37 & 0.76 & 0.97 \\
    \textbf{Base-OM (p1)} & 0.45 & 0.82 & 0.97 \\
    \textbf{Base-OM (all)}  & 0.56 & \textbf{0.92} & \textbf{0.99} \\
    \textbf{LLM-PE (p1)} & 0.40 & 0.75 & 0.97 \\
    \textbf{LLM-PE (all)} & 0.32 & 0.69 & 0.93 \\
    \bottomrule
  \end{tabular}
\end{table}

\begin{table}[t]
  \centering
  % \scriptsize
  \setlength{\tabcolsep}{1.5pt}
  \caption{
    Preference estimation error for each opponent agent by $p_1$.
    Values represent the Mean Squared Error (MSE) between the estimated and true scores ($s_n^m$).
    The "Avg" column shows the overall estimation accuracy across all agents. Bold values indicate the minimum error.}
  \label{tab:accuracy_game_a}
  \begin{tabular}{lcccccc}
    \toprule
    \textbf{Method} & \textbf{Mayor} & \textbf{Cities} & \textbf{Union} & \textbf{DoT} & \textbf{Env} & \textbf{Avg} \\
    \midrule
    \textbf{Proposed} & 159 & \textbf{217} & \textbf{120} & 99 & 201 & \textbf{159} \\
    \textbf{Base-OM} & \textbf{112} & 232 & 155 & 120 & 324 & 189 \\
    \textbf{LLM-PE} & 167 & 238 & 185 & \textbf{96} & \textbf{129} & 163 \\
    \bottomrule
  \end{tabular}
\end{table}

\subsection{Results and Discussion}
In this section, we evaluate the negotiation outcomes and preference estimation accuracy. Table~\ref{tab:outcomes_game_a} and Table~\ref{tab:accuracy_game_a} present these results.

\paragraph{Analysis of Negotiation Outcomes.}
The proposed method (\textit{all}) achieved the strongest overall performance on FAR, while maintaining a competitively high PAR. In particular, it obtained the highest FAR (0.62), indicating a strong capability to identify agreements acceptable to all agents under complex multi-agent interactions. This result highlights the advantage of explicitly modeling preferences using linguistic signals extracted by LLMs. 

Compared to the single-estimator setting (\textit{p1}), mutual preference estimation (\textit{all}) further improved both FAR and PAR, demonstrating enhanced strategic coordination among agents.

\paragraph{Analysis of Estimation Accuracy.}
The proposed method achieved a lower Mean Squared Error (MSE) ($159$) than the Base-OM approach ($189$), showing that natural language information improves preference estimation. 
Although its average error was slightly lower (i.e., better) than that of LLM-PE, the proposed method showed more balanced accuracy across agents, enabling less biased preference prediction. 
This likely helped the agents propose deals $d_t$ that better satisfy the complex multi-party constraints needed for consensus.

\section{Conclusion}
In this paper, we address the challenge of quantitatively modeling opponent preferences from natural language utterances in negotiation. We propose a novel Bayesian preference estimation framework that integrates numerical proposals with qualitative natural language signals extracted from dialogue. Our experiments in a multi-agent, multi-issue negotiation setting demonstrated that the proposed method achieves consistently higher FAR than baselines relying solely on numerical data or direct LLM inference, while also maintaining a high PAR. These findings highlight the potential of combining the linguistic intelligence of LLMs with mathematically rigorous Bayesian inference to facilitate effective conflict resolution.

\section*{Limitations}
While our framework improves multi-party negotiation, there remain several avenues for future enhancement. 
First, although we validated our method using a complex benchmark, further research is required to verify its generalizability across more diverse utility structures and larger agent populations. 
Second, while we achieve high accuracy by assuming sincere dialogue, robustness could be improved by incorporating mechanisms to account for strategic behaviors like deception or bluffing. 
Third, while we focused on learning preference shapes, integrating the inference of opponents' reservation values would allow for more sophisticated coordination, especially in ambiguous settings where agreement zones are difficult to identify.
Finally, the computational complexity increases as the number of issues and options grows, although this challenge may be mitigated by incorporating approximation algorithms from prior research on opponent modeling.

\section*{Acknowledgments}

This work was supported by a joint research project with Accenture Japan Ltd (150241400037), JSPS KAKENHI Grants (JP23K11230, JP22H00533), and JST CREST (JPMJCR20D1).
We also thank Gakuse Hoshina, Kaori Fujiwara, and Atsushi Suyama for their valuable support and feedback.

\bibliography{custom}

\clearpage
\appendix

\section{Negotiation Scenario Details}
\label{sec:appendix_scenario}

To enable full reproduction of our experiments and independent verification of our results, 
we provide the complete specification of the Harbour Sport Park negotiation scenario, including party roles, issue definitions, and preference profiles.

\subsection{Parties and Roles}
Table~\ref{tab:party_roles} describes the six stakeholders involved in the Harbour Sport Park negotiation and their strategic characteristics.

\begin{table*}[htbp]
\centering
\caption{Roles and characteristics of negotiation parties.}
\label{tab:party_roles}
\begin{tabular}{lp{10cm}}
\toprule
\textbf{Party} & \textbf{Characteristics} \\
\midrule
SportCo & Proposer and facilitator of the project. Holds veto power over the final decision. \\
Department of Tourism; DoT & Provider of federal funding. Holds veto power over the final decision. \\
Environmental League; Env & Prioritizes ecological preservation above all else. \\
Local Labour Union; Union & Advocates for union priority in employment rules. \\
Other Cities; Cities & Demands increased compensation for neighboring municipalities. \\
Mayor & Participates as the head of the host city. \\
\bottomrule
\end{tabular}
\end{table*}

\subsection{Issues and Options}
The negotiation consists of five issues, each with three to five options. Table~\ref{tab:issue_options_clean} provides the detailed definitions for each option.

\begin{table*}[htbp]
\centering
\caption{Definitions of negotiation issues and options.}
\label{tab:issue_options_clean}
\begin{tabular}{lccccc}
\toprule
\textbf{Issue} & \textbf{Option 1} & \textbf{Option 2} & \textbf{Option 3} & \textbf{Option 4} & \textbf{Option 5} \\
\midrule
A: Infrastructure & Water-based & Amphibious & Land-based & - & - \\
B: Ecology & Accept damage & Balanced & Max effort & - & - \\
C: Employment & Union priority & 2:1 Ratio & 1:1 Ratio & No priority & - \\
D: Fed. Funding & \$3B & \$2B & \$1B & None & - \\
E: Compensation & \$600M & \$450M & \$300M & \$150M & None \\
\bottomrule
\end{tabular}
\end{table*}

\subsection{Preference Profiles}
Table~\ref{tab:agent_scores} presents the private score functions and reservation thresholds ($\tau$) for each party. Under these high thresholds, the negotiation is highly challenging, with only 0.4\% of all possible deals satisfying the conditions for a full agreement.

\begin{table*}[htbp]
\centering
\setlength{\tabcolsep}{3pt}
\caption{Agent score functions and reservation thresholds ($\tau$).}
\label{tab:agent_scores}
\begin{adjustbox}{max width=\textwidth}
\begin{tabular}{lcccccc}
\toprule
\textbf{Party} & \textbf{Threshold ($\tau$)} & \textbf{Issue A} & \textbf{Issue B} & \textbf{Issue C} & \textbf{Issue D} & \textbf{Issue E} \\
\midrule
SportCo (Veto) & 53 & [14, 8, 0] & [11, 7, 0] & [0, 5, 10, 17] & [35, 29, 20, 0] & [0, 5, 10, 15, 23] \\
Department of Tourism; DoT (Veto) & 70 & [0, 11, 5] & [0, 20, 25] & [0, 2, 4, 9] & [10, 26, 40, 0] & [4, 8, 15, 12, 0] \\
Environmental League; Env & 45 & [0, 22, 45] & [0, 25, 55] & [0, 0, 0, 0] & [0, 0, 0, 0] & [0, 0, 0, 0, 0] \\
Local Labour Union; Union & 50 & [15, 20, 0] & [0, 0, 0] & [42, 35, 25, 0] & [30, 20, 10, 0] & [2, 4, 6, 8, 0] \\
Other Cities; Cities & 50 & [0, 4, 10] & [0, 0, 0] & [12, 8, 6, 0] & [0, 8, 13, 18] & [60, 45, 30, 15, 0] \\
Mayor & 55 & [14, 8, 0] & [12, 8, 0] & [24, 18, 12, 0] & [40, 30, 23, 0] & [0, 2, 4, 7, 10] \\
\bottomrule
\end{tabular}
\end{adjustbox}
\end{table*}

\section{Experimental Setup and Computational Resources}
\label{sec:appendix_experiment}

We describe the implementation details and hyperparameter 
settings to support reproducibility of our experiments.

\subsection{Implementation and Infrastructure}
The proposed framework was implemented by extending the open-source negotiation environment provided by \citet{cooperation}, which is distributed under the MIT License. All experiments were conducted within a Docker container on an Amazon Web Services (AWS) instance, accessed via SSH from a macOS workstation. The inference process utilized the GPT-4.1 model via the OpenAI API.
Total execution time for 500 trials of the proposed method was several hours.

\subsection{Bayesian Hypothesis Space and Parameters}
Following the methodology in \cite{opponent-modeling}, the hypothesis space $H$ is defined as follows:
\begin{itemize}
    \item \textbf{Issue Weights ($\mathbf{w}$):} We consider all possible permutations of issue rankings. For $M=5$ issues, this results in $5! = 120$ weight hypotheses.
    \item \textbf{Evaluation Functions ($\mathbf{v}$):} Preference shapes are modeled as linear functions. Given the issues' options (A, B: 3; C, D: 4; E: 5), the total number of combinations of evaluation functions is $3 \times 3 \times 4 \times 4 \times 5 = 720$ patterns.
    \item \textbf{Likelihood Parameters:} The standard deviation for numerical offer likelihood calculation was set to $\sigma = 1.0$.
    \item \textbf{LLM Settings:} The sampling temperature for the LLM was set to $0$ to ensure deterministic reasoning, with \texttt{max\_tokens} maintained at the default configuration of the GPT-4.1 model.
\end{itemize}

\section{Societal Impact}
\label{sec:appendix_impact}
The primary goal of this research is to enhance the transparency and efficiency of consensus-building. By integrating LLMs with structured Bayesian inference, we provide an interpretable framework where an agent's internal belief state can be inspected. This facilitates smoother decision-making in multi-party organizational contexts. We acknowledge potential risks: while our framework assumes truthful signaling, future iterations could be misused for deceptive strategy optimization. We recommend human oversight in high-stakes deployments.

\section{Use of AI Assistants}
\label{sec:appendix_ai}
To ensure transparency regarding the use of AI assistance, we explicitly specify the stages of the research in which AI tools were utilized. Large Language Models were employed to assist in (1) literature research and information synthesis, (2) debugging and optimizing parts of the experimental code implementation, and (3) refining and proofreading the manuscript to improve clarity and grammatical accuracy. All final decisions, scientific interpretations, and content verifications were performed by the authors.

\section{Prompt for Signal Extraction}
\label{sec:appendix_prompt}

We provide the full prompt used in our signal extraction 
module to facilitate reproducibility.

Figure~\ref{fig:signal_extraction_prompt} illustrates the detailed prompt used to extract qualitative signals $z_t$ from negotiation dialogues.

\begin{figure*}[htbp]
    % \footnotesize
    \centering
    \begin{tcolorbox}[colback=white, colframe=black, sharp corners, boxrule=0.5pt]
        You are an expert in negotiation analysis. 
        Analyze the following chat history and extract opponent modeling signals for each agent.
        
        \vspace{1em}
        \textbf{Negotiation Rules:}\\
        \{negotiation\_rule\}
        
        \vspace{1em}
        \textbf{Chat history:}\\
        \{chat\_history\}

        \vspace{1em}
        \textbf{Task:}\\
        For each agent appearing in the chat history, extract structured opponent modeling signals. Identify behavioral signals that indicate their preferences.

        \vspace{1em}
        \textbf{Important instructions:}
        \begin{enumerate}
            \item Be sure to extract at least one signal for each agent.
            \item Include signals that can be inferred by comprehensively considering the chat history and negotiation rules, even if the agent did not directly mention them in their statement. Do not limit signal extraction only to the options proposed in the deal; also extract signals regarding issue preferences and comparisons of preferences between two issues or options.
            \item Extract signals in chronological order as they appear in the chat history. Process the conversation from beginning to end, and add signals to the array in the order you encounter them.
            \item Classify each signal using the following information:
        \end{enumerate}

        \renewcommand{\labelitemi}{-}
        \renewcommand{\labelitemii}{-}

        \begin{itemize}
            \item \texttt{entity}: The type of reference ("issue" or "option")
            \begin{itemize} 
                \item "issue": Refers to the name of an issue (e.g., A, B)
                \item "option": Refers to a specific choice within an issue (e.g., A1, B1)
            \end{itemize}
            
            \item \texttt{signal\_type}: The type of signal ("point" or "comparison")
            \begin{itemize}
                \item "point": A direct preference toward a specific target ("A", "A1")
                \item "comparison": A preference comparison between two targets ("A, B", "A1, B1")
            \end{itemize}
            
            \item \texttt{target}: The specific target (e.g., "A", "A1", "A, B", "A1, B1")
            
            \item \texttt{stance}: The agent’s position toward the target ("prefer" or "oppose"). When the agent gives importance to the target, use "prefer". When the agent devalues or rejects the target, use "oppose".
        \end{itemize}

        \vspace{1em}
        When extracting signals, pay particular attention to the stated reasons why the agent proposed a specific deal. Accurate capture of each party’s preferences by considering their public answers and the flow of proposed deals is required. While the agent is likely to prefer the option they proposed, consider that they may have compromised to accommodate other parties; thus, an even more preferred option might exist.

        \vspace{1em}
        Using the provided function schema, create a structured response with agent names as keys. Do not return empty results. Always extract at least some signals.
    \end{tcolorbox}
    \caption{The prompt for the qualitative signal extraction.}
    \label{fig:signal_extraction_prompt}
\end{figure*}

\end{document}